# Mainumby: un Ayudante para la Traducción Castellano-Guaraní

Michael Gasser

*Escuela de Información, Computación, y Ingeniería*
*Universidad de Indiana, Estados Unidos de América*

7 agosto, 2018

## Introducción

Las computadoras y una amplia gama de aplicaciones juegan un papel importante en el trabajo cotidiano del traductor moderno. Sin embargo, las herramientas computacionales diseñadas para ayudar en el proceso de traducción benefician solamente la traducción desde o hacia una pequeña minoría de los 7.000 idiomas del mundo, los que podemos llamar "idiomas privilegiados", el inglés, el castellano, el alemán, el chino, el árabe, etc. En cuanto a los traductores que trabajan con los demás idiomas, los idiomas que quedan minorizados en el mundo digital, no se pueden aprovechar de las herramientas que están acelerando la producción de traducciones en los idiomas privilegiados. Cabe preguntar si es posible colmar la brecha entre lo que está disponible para esos idiomas y para los minorizados. Esta ponencia propone un marco para la traducción asistida por computadora hacia idiomas minorizados y su implementación en una aplicación para la traducción castellano-guaraní.

Primero, se resumen las herramientas computacionales existentes para la traducción y el recurso principal del que dependen, los corpus bilingües. Segundo, se considera lo que es posible sin un corpus adecuado, y se presenta un marco para la traducción asistida por computadora basada en reglas en vez de un corpus. Finalmente, se describe Mainumby, una implementación en desarrollo de este marco para la traducción del castellano al guaraní.

## Herramientas y recursos computacionales para la traducción

### Corpus multilingües y memorias de traducción

El recurso más importante para el uso de la tecnología en la traducción es un **corpus multilingüe**, que consiste en documentos en dos o más lenguas. Como el sistema que usa el corpus debe buscar coincidencias en el corpus con una oración fuente arbitraria, cuánto más amplio sea el corpus, más útil será. Corpus bilingües amplios están dispuestos para los idiomas privilegiados en el mundo digital, como el castellano, el inglés, y el chino, pero no para idiomas minorizados como el guaraní.

Para ser utilizable, un corpus bilingüe se convierte en una **memoria de traducción**. Se limpia y después se alinean las oraciones de los documentos en el corpus. Es decir, cada unidad o segmento en la memoria de traducción que resulta consiste en oraciones que son traducciones el uno del otro. Las memorias de traducción se aplican en dos tipos de sistemas, los de la traducción asistida por computadora y los de la traducción automática.

Un ejemplo de una memoria de traducción en el dominio de la medicina es una recopilada por el Centro Europeo para la Prevención y el Control de las Enfermedades. Consiste en aproximadamente 3.000 oraciones ("unidades de traducción") en 25 idiomas europeos. En la Figura 1 se ven los "segmentos" (entre "**\<seg>**" y "**\</seg>**") ingleses y castellanos de una de esas unidades (entre "**\<tu>**" y "**\</tu>**").

```
<tu>
<tuv xml:lang="EN">
<seg>Less than 0.5% of those who get the infection die.</seg>
</tuv>
<tuv xml:lang="ES">
<seg>Fallecen menos del 0,5 % de las personas que contraen esta infección.</seg>
</tuv>
</tu>
```

Figura 1. Unidad de traducción de la memoria de traducción del CEPCE

*La traducción asistida por computadora*

De manera general, la traducción asistida por computadora (TAC) consiste en aplicaciones basadas en memorias de traducción para ayudar a traductores en su trabajo. El usuario del un sistema TAC normalmente elige las memorias que se incluyen en la aplicación en el tiempo de traducción, a menudo restringidas a memorias con traducciones hechas por traductores fiables y tratando el dominio del texto fuente.

Para ilustración, veamos el rendimiento de OmegaT (Smolej, 2016), una aplicación TAC gratuita y de código abierto. En este ejemplo, se ha cargado la memoria de traducción del CEPCE descrita arriba. El documento para traducir consiste en cuatro oraciones que se parecen en diferentes grados a las oraciones en el corpus mismo (es decir, en la memoria de traducción). En la Figura 2, se muestra la interfaz de OmegaT y su rendimiento cuando se ha seleccionado la tercera oración fuente en el documento, la que se parece a la oración en la Figura 1.

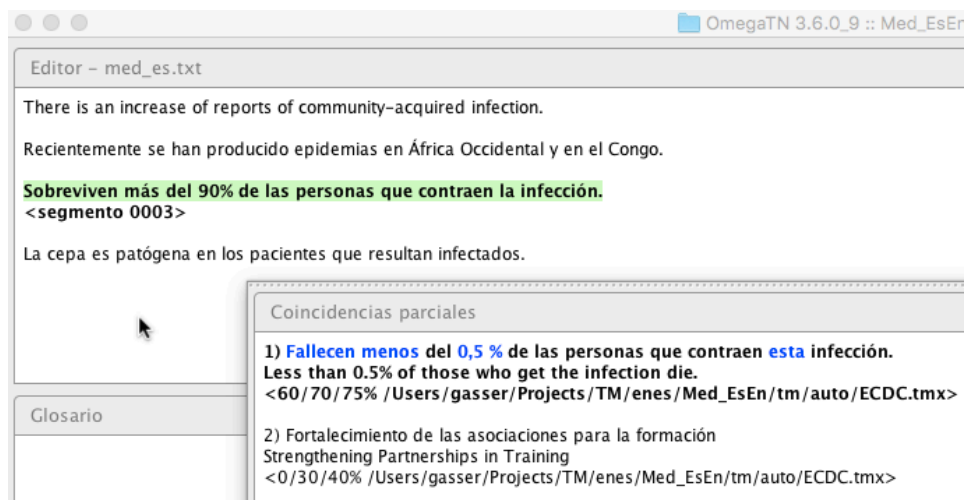

Figura 2. Interfaz de OmegaT durante la traducción de un documento.

La aplicación muestra en su ventana de coincidencias las oraciones en la memoria de traducción cargada que coinciden mejor a la oración fuente. Como la coincidencia en este caso no es perfecta, el usuario debe descubrir cuáles son los segmentos de la traducción inglesa que puede mantener en su traducción y cuáles son los que debe reemplazar.

A medida que el traductor usa una aplicación TAC, el sistema almacena sus traducciones en una memoria de traducción que se puede guardar de manera separada o integrar en una memoria existente. Cuando el traductor traduzca un documento parecido, se beneficia de esta memoria de traducción guardada.

*La traducción automática*

Otra aplicación de las memorias de traducción (aunque normalmente no se llaman "memorias de traducción" en este campo) es la traducción automática (TA). Se trata de sistemas que producen traducciones completas de oraciones fuente.

El desarrollo de un sistema de TA incluye una fase de aprendizaje, durante la que el sistema usa la memoria de traducción para aprender correspondencias entre palabras o frases en los dos idiomas. Al tiempo de traducción, el sistema busca en su memoria de correspondencias coincidencias con segmentos de la oración fuente, genera las traducciones de los segmentos, y las ordenan par producir la oración meta.

El traductor de Google (Wu et al., 2016) es un sistema de TA muy conocido. Actualmente traduce entre cada par de idiomas en una lista de 104. Para pares de lenguas que se benefician de corpus substanciales, como el inglés y el castellano, los resultados son bastante impresionantes. Esto se debe primero a la obsesión de Google con la recopilación de datos, en este caso ejemplos de traducciones entre el inglés y el castellano encontrados en Internet, y segundo a la introducción de métodos neuronales muy potentes en 2016.

Como ilustración, se dieron al traductor de Google las mismas oraciones que se habían presentado a OmegaT en el ejemplo anterior, y las tradujo todas al inglés perfectamente. Por ejemplo, la oración castellana resaltada en la Figura 2 se traduce como *More than 90% of people who contract the infection survive.*

A pesar de este rendimiento impresionante, cabe tener en cuenta que no podemos depender de traducciones perfectas en la mayoría de los casos; ciertamente seguiremos necesitando a traductores humanos para todos los pares de idiomas.

*El desafío de las lenguas sin muchos recursos*

Para traducir desde o hacia idiomas que pertenecen al pequeño grupo de los privilegiados, no existen corpus bilingües suficientes para crear memorias de traducción útiles. Un alternativa temporal es un sistema de **traducción automática basada en reglas**, el que se puede implementar utilizando recursos más disponibles: diccionarios digitales (o digitalizables), analizadores y generadores morfológicos (los que se pueden implementar con descripciones gramaticales), y, si existen, corpus monolingües. En la siguiente sección, se describe tal sistema.

## La Traducción por Segmentos Generalizados y Mainumby

*La Traducción por Segmentos Generalizados*

La Traducción por Segmentos Generalizados (TSG, también llamada Traducción de Dependencias Minimales) es una teoría de la traducción automática basada en reglas y un marco para la traducción asistida por computadora que incorpora la teoría (Gasser, 2017). El proyecto TSG tiene como objetivos:

- Ayudar a traductores a traducir documentos de un idioma privilegiado (IP) a un idioma minorizado (IM).
- Crear (o aumentar) memorias de traducción del IP al IM.

TSG realiza la traducción de segmentos, no de oraciones enteras. Ofrece sugerencias múltiples para la traducción de cada segmento de una oración traducida y la posibilidad de rechazar todas las sugerencias e introducir una traducción totalmente diferente. Registra las traducciones hechas por usuarios en memorias de traducción.

Los recursos necesarios para la implementación de TSG para un par de idiomas incluyen un diccionario digital, un analizador morfológico del idioma fuente, un generador morfológico del idioma meta, y una descripción de las mayores diferencias gramaticales de los idiomas.

*Mainumby*

Mainumby es una implementación en desarrollo del marco TSG para la traducción del castellano al guaraní. Se basa en recursos que estaban disponibles en el año 2015 al principio del proyecto.[1]

*Ejemplos del rendimiento*

Consideremos cómo el sistema traduce frases. Los casos más simples son frases fijas como *por eso*. Mainumby traduce una frase fija castellana a una frase (o palabra) fija en guaraní, ofreciendo opciones si hay múltiples traducciones (Figura 3).[2]

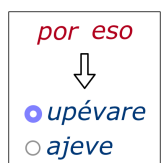

Figura 3. Traducción de una frase fija.

Más complejas son palabras o frases que involucran la morfología, es decir, en la traducción de sustantivos, adjetivos, y verbos. La Figura 4 muestra en ejemplo, la traducción de la palabra *llamo*. En esos casos, Mainumby analiza la palabra castellana, generando una raíz y unas características gramaticales, como el tiempo y las características del sujeto del verbo (persona=1, número=plural en este caso). Entonces se traduce la raíz y finalmente se genera la palabra guaraní de la raíz guaraní y las características gramaticales copiadas de la palabra castellana.[3]

---

[1] Ciertamente, mucho otros recursos se han hecho dispuestos desde entonces; sobre todo, las
[2] En las figuras, rojo representa castellano, azul guaraní.
[3] Los caracteres "<h" representan la consonante triforme (Krivoshein de Canese & Acosta Alcaraz, 2007), que se realiza como carácter pronunciable después de la generación.

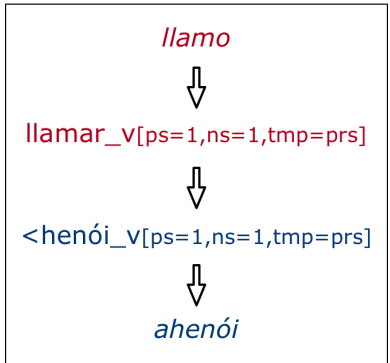

Figura 4. Traducción de una palabra con morfología compleja.

Aún más complejas son frases que contienen palabras gramaticales castellanas que no corresponden a palabras guaraníes sino a morfemas. La Figure 5 muestra un ejemplo, la traducción de la frase *no te llamo*. En esos casos se aplica una o más transformaciones morfosintácticas a la frase, por las que se guaraniza parcialmente el castellano. En este ejemplo las palabras *no* y *te* se eliminan y se incorporan como características gramaticales adjuntadas a la raíz, es decir, como negación y objeto de segunda persona singular.

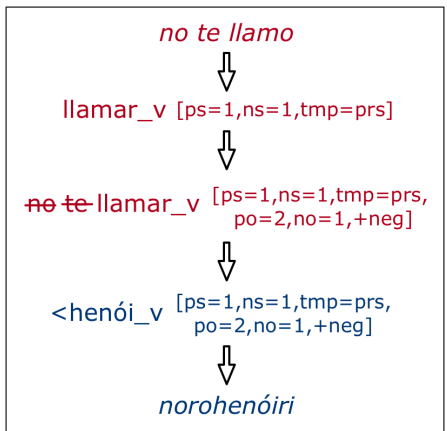

Figura 5. Traducción de palabras gramaticales.

El sistema puede también traducir frases más complejas que incluyen dos palabras léxicas y también correspondencias sintácticas entre las lenguas. La Figura 6 muestra un ejemplo, la traducción de la frase *se burla del presidente*. Después del análisis morfológico del verbo *burla* y del sustantivo *presidente*, una transformación morfosintáctica reemplaza el pronombre *se* con la característica +rflx,[4] así distinguiendo el verbo transitivo *burlar* del verbo pronominal *burlarse*. El léxico tiene una entrada que asocia el verbo *burlarse* (burlar_v[+rflx]) con el verbo guaraní *ñembohory*. Esta entrada tiene también el conocimiento de que "burlarse de $s" en castellano corresponde a "ñembohory $s rehe" en guaraní, donde "$s" representa cualquier sustantivo. Para *presidente*, Mainumby encuentra otra entrada en su léxico, que asocia este sustantivo (más bien, su raíz) con el sustantivo guaraní *mburuvicha*. Las dos entradas se funden,

---

[4] Mainumby no distingue los verbos reflexivos y los verbos pronominales.

convirtiéndose en un segmento que se realiza, después de la generación morfológica, como *oñembohory mburuvicha rehe*.

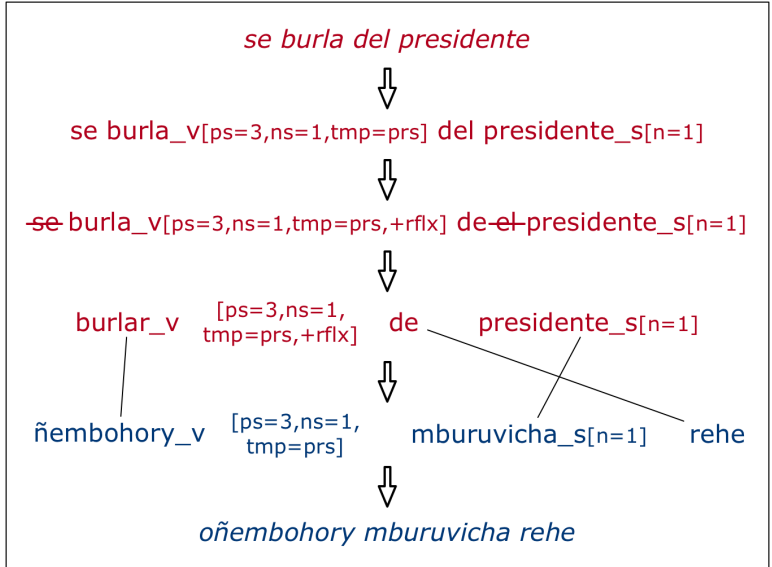

Figura 6. Traducción de frases con dos palabras léxicas.

*Interfaz*

En la Figura 7, se ve la interfaz de Mainumby donde un usuario introduce oraciones para traducir. En la actual versión de la aplicación, o escribe directamente en el espacio o copia y pega el texto de otro documento. En futuras versiones, se podrá subir documentos para traducir de la computadora del usuario.

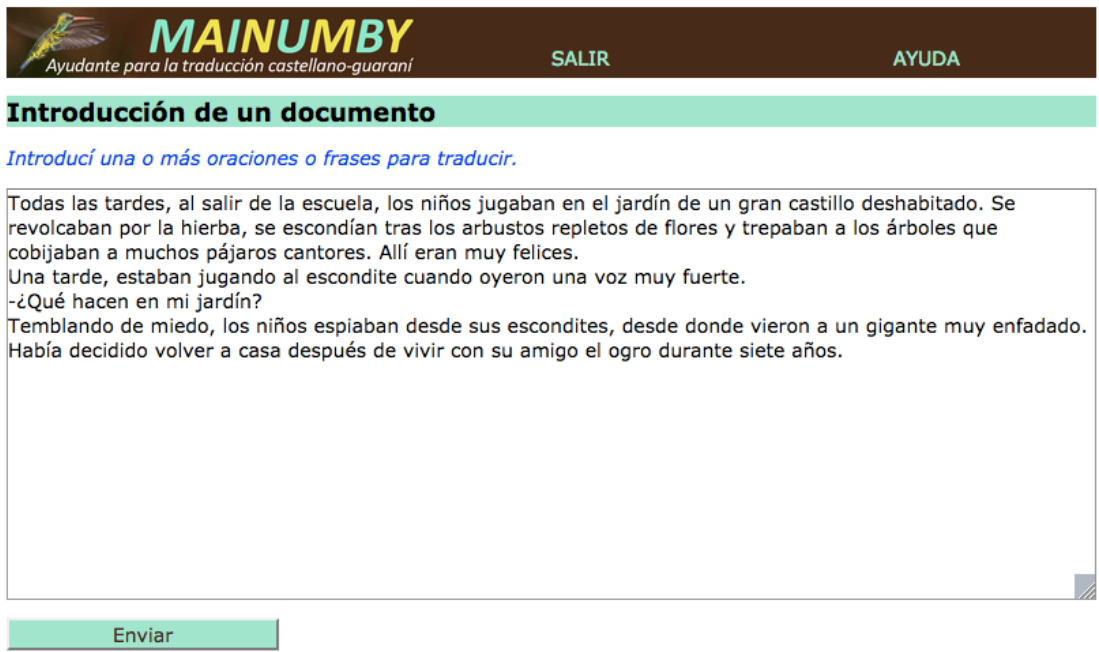

Figura 7. Interfaz de Mainumby para introducir documentos.

En la Figura 8, se ve la ventana en la que el usuario selecciona o introducir su traducción de una oración. En el espacio "Actual oración", se muestra la oración fuente en segmentos coloreados. Los grises representan partes de la oración que Mainumby no pudo traducir. Cuando el usuario hace clic en un segmento de la oración fuente, las opciones que Mainumby ofrece para traducir el segmento se muestran en el espacio "Frase", como ilustrado en la figura para el segmento *por la hierba*. Entonces el usuario puede seleccionar una de las opciones o también rechazar todas y introducir su propia traducción en el espacio provisto. Cada oración que el usuario traduce con (o sin) ayuda de Mainumby se incluye en el espacio "Documento". Finalmente, la traducción del documento se puede guardar en un archivo en la computadora del usuario.

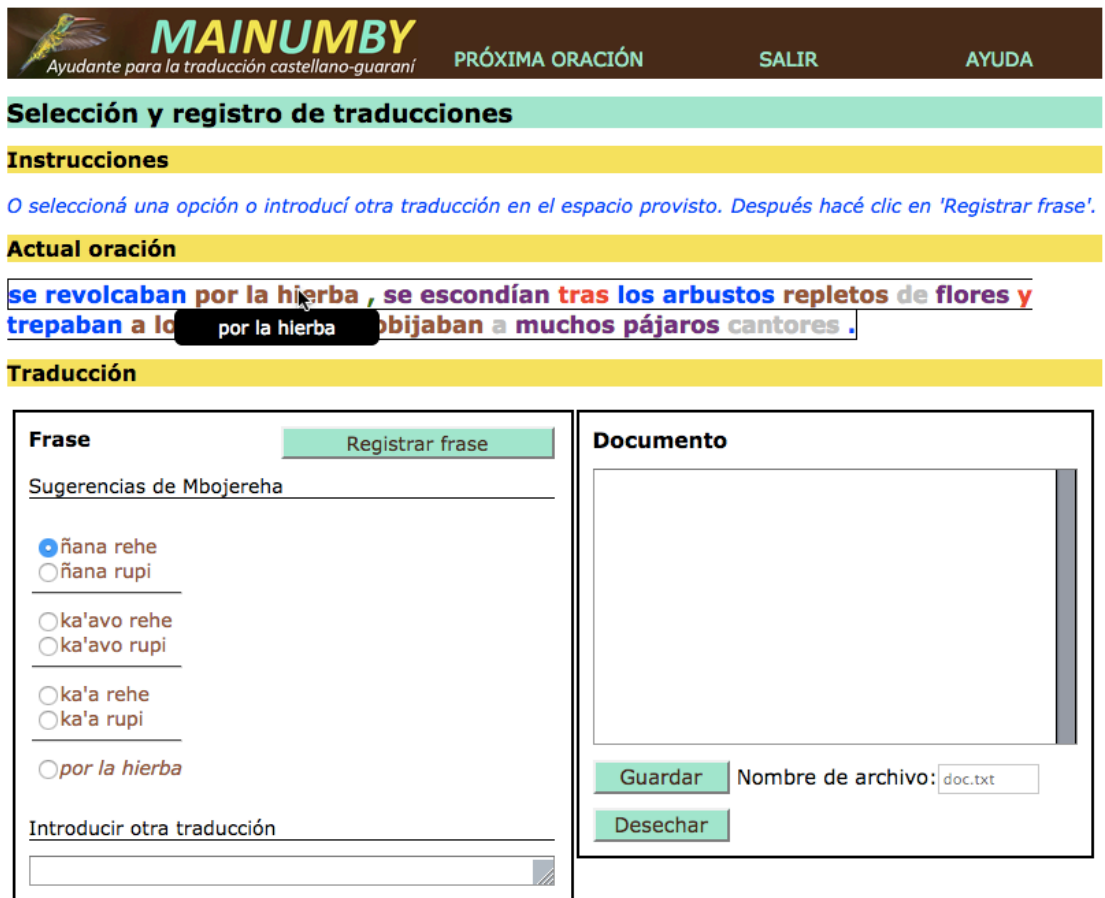

Figura 8. Interfaz de Mainumby para selección de traducciones de segmentos.

Las traducciones hechas por usuarios se registran, formando parte de las memorias de traducción que Mainumby construye a medida que se utiliza.

*Componentes y procesamiento de Mainumby*

Mainumby consiste en un léxico, un conjunto de reglas de transformación, un analizador morfológico del castellano, un generador morfológico del guaraní, y el algoritmo que realiza el procesamiento de traducción.

El componente central del sistema es el léxico. Las entradas en el léxico son **segmentos generalizados** (SG); a través de ellos se realizan traducciones de palabras, frases fijas, raíces, y frases más complejas, y también reglas de concordancia. Por ejemplo, el SG mostrado en la Figura 9 contiene reglas de concordancia estipulando que el sujeto, el tiempo, y la negación del verbo se copie del castellano al guaraní. De esta manera, este SG se puede aplicar no solo a la traducción de la frase mostrada en la Figura 6, *se burla del presidente*, sino también a miles de otros posibles frases como *me burlaba del profesor* y *no se van a burlar de nadie*. Es en este sentido que un SG es "generalizado".

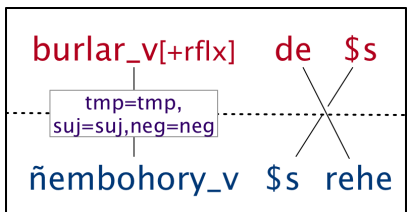

Figura 9. Segmento generalizado para la traducción de *burlarse de alguien*.

Otro componente consiste en **reglas de transformación morfosintáctica** a través de las que frases castellanas se guaranizan en palabras guaraníes. La Figura 10 muestra un ejemplo, una regla que busca una secuencia consistiendo en *no* y cualquier verbo, anula la palabra *no*, y agrega la característica +neg al verbo.

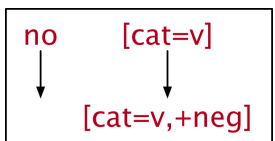

Figura 10. Regla de transformación morfosintáctica para verbos negativos.

Finalmente el sistema depende de un **analizador morfológico** de verbos, sustantivos, y adjetivos castellanos y un **generador morfológico** de verbos, sustantivos, y adjetivos guaraníes. Ellos se han implementado en el marco bien establecido en la morfología computacional de transductores de estado finito (Gasser, 2009). La Figura 11 ilustra un fragmento del generador morfológico de verbos guaraníes. Estipula que un verbo de la subcategoría *areal* con sujeto de primera persona plural se realice con un prefijo antes de la raíz (R), o el de la forma inclusiva (*<j>a-*, realizado como *ja-* o *ña-*) o el de la forma exclusiva (*ro-*).

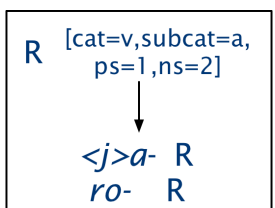

Figura 11. Fragmento del generador morfológico de verbos guaraníes: sujeto primera persona plural.

El resto del sistema consiste en el código informático que implementa el procesamiento. Este componente es independiente de los idiomas; es decir, se puede aplicar también a otros pares de idiomas si los recursos necesarios están dispuestos para esos idiomas. Ya existe una implementación para la traducción desde el inglés hacia el amárico (Gasser, 2017).

La traducción de una oración fuente sigue estos pasos:

Fuente

1. Análisis léxico y morfológico de la oración
2. Transformaciones morfosintácticas que coinciden con la oración
3. Búsqueda de segmentos generalizados en el léxico que coinciden con la oración

Traducción léxica y sintáctica

4. Selección, instanciación, y fusión de segmentos generalizados coincidentes

Meta

5. Realización de reglas de concordancia
6. Ordenación de palabras dentro de segmentos
7. Generación morfológica de palabras

*Estatus del proyecto y pasos futuros*

Actualmente Mainumby tiene un léxico de aproximadamente 10.000 segmentos generalizados, la mayoría extraídos de un diccionario bilingüe digital.[5] Tiene además un conjunto de aproximadamente 100 reglas de transformación morfosintácticas. Los componentes más amplios de la aplicación son el analizador y el generador morfológicos, compilados en tres ficheros, de 0,5 MB, 3 MB, y 12 MB. Las reglas y procesadores morfosintácticos se basan en descripciones publicadas de la gramática guaraní (Ayala, 1996; Krivoshein de Canese & Acosta Alcaraz, 2007; Zarratea, 2002).

La implementación es de código abierto, y todo el código informático y los datos están descargables a través de este enlace: https://github.com/hltdi/mainumby. Mainumby se ha implementado en forma de una aplicación web, accesible a través de este enlace: http://plogs.soic.indiana.edu/mainumby.

El sistema se podrá evaluar solamente con la participación de traductores. Informalmente, las fortalezas de Mainumby incluyen la traducción de la morfología y la facilidad de agregar entradas (segmentos generalizados) al léxico que permiten la traducción de muchas frases. Sus debilidades incluyen grandes lagunas en su conocimiento léxico (debido a las limitaciones del diccionario en el que se basa) y la brevedad de los segmentos en la traducciones (véase la oración en la Figura 8 para un ejemplo). Ambas se podrían superar hasta cierto punto por la inclusión de más información léxica proveída por usuarios del sistema.

La siguiente fase del proyecto empezará con la participación de traductores del castellano al guaraní experimentados. Incluirá:

1. Evaluación del marco lingüístico y de la interfaz por traductores y mejoramiento según los resultados
2. Aumento y mejoramiento a través de nuevos recursos disponibles (diccionarios, corpus)

---

[5] Desafortunadamente, no ha sido posible determinar el autor de este diccionario.

3. Implementación de métodos para aprender de las traducciones y correcciones de usuarios

La fase final del proyecto consistirá en:

1. Lanzamiento como herramienta para la comunidad de traductores (y tal vez estudiantes del guaraní)
2. Lanzamiento de memorias de traducción que resultan para inclusión en sistemas TAC y TA

## Conclusiones

Sistemas de la traducción automática (TA) y la traducción asistida por computadora (TAC) convencionales dependen de corpus bilingües grandes, elaborados como memorias de traducción (MT). Hasta que MT adecuadas estén dispuestas, puede ser útil un sistema de TA/TAC rudimentario, basado en diccionarios y implementaciones computacionales de descripciones de la gramática básica de los idiomas y sus diferencias. Tal sistema podría facilitar tanto la traducción humana como la recopilación de MT.

La Traducción por Segmentos Generalizados es un intento inicial hacia un marco para desarrollar ese tipo de sistema con una implementación experimental (Mainumby) para la traducción castellano-guaraní. Sin embargo, todavía queda el paso esencial de determinar si Mainumby, o una aplicación parecida, sirva la comunidad de traductores paraguayos como fue diseñado para hacer. Es decir, de ahora en adelante, el trabajo debe ser colaborativo, incorporando las experiencias de traductores en el diseño de la interfaz, el mejoramiento del conocimiento del sistema, y la recopilación de memorias de traducción.

## Bibliografía